\documentclass[sigconf]{acmart}
\usepackage{siunitx}

\AtBeginDocument{%
  }

\acmYear{2025}

\acmConference[2nd HEAL Workshop at CHI'25, April 26-May 1]{}{2025}{Yokohama, Japan}

\begin{document}

\title{Name of Thrones: Evaluating How LLMs Rank Student Names, Race, and Gender in Status Hierarchies}

\author{Annabella Sakunkoo}
\authornote{Both authors contributed equally to this research.}
\email{apianist@ohs.stanford.edu}
\orcid{}
\affiliation{%
  \institution{Stanford University OHS}
  \state{California}
  \country{USA}
}

\author{Jonathan Sakunkoo}
\email{jonkoo@ohs.stanford.edu}
\authornotemark[1]
\affiliation{%
  \institution{Stanford University OHS}
  \state{California}
  \country{USA}
}

\renewcommand{\shortauthors}{Sakunkoo and Sakunkoo}

\begin{abstract}
Across cultures, names tell a lot about their bearers as they carry deep personal, historical, and cultural significance. Names have also been found to serve as powerful signals of gender, race, and status in the social hierarchy--a pecking order in which individual positions shape others’ expectations on their perceived competence and worth \cite{podolny2005}. With the widespread adoption of Large Language Models (LLMs) and given that names are often an input for LLMs, it is crucial to evaluate whether LLMs may sort people into status positions based on first and last names and, if so, whether it is in an unfair, biased fashion. While prior work has primarily investigated biases in first names, little attention has been paid to last names and even less to the combined effects of first and last names. In this study, we conduct a large-scale analysis with bootstrap standard errors of 45,000 name variations across 5 ethnicities to examine how AI-generated responses exhibit systemic name biases. Our study investigates three key characteristics of inequality and finds that LLMs reflect, construct, and reinforce status hierarchies based on names that signal gender and ethnicity as they encode differential expectations of competence, leadership, and economic potential. Contrary to the common assumption that AI tends to favor Whites, we show that East and, in some contexts, South Asian names receive higher rankings. We also disaggregate Asians, a population projected to be the largest immigrant group in the U.S. by 2055 \cite{pewresearch_modern_2015}. Our results challenge the monolithic Asian model minority assumption, illustrating a more complex and stratified model of bias. Gender moderates biases, with girls having advantages in certain racial groups while facing unfair disadvantages in others. Additionally, spanning cultural categories by adopting Western first names improves AI-perceived status for East and Southeast Asian students, particularly for girls. Our findings underscore the importance of intersectional and more nuanced understandings of race, gender, and mixed identities in the evaluation of LLMs, rather than relying on broad, monolithic, and mutually exclusive racial or gender categories. By examining LLM bias and discrimination in our multicultural educational contexts through the sociological status lenses, our study illustrates potential harms of using LLMs in education as they do not merely reflect implicit biases but also actively construct new social hierarchies that can unfairly shape long-term life trajectories.

\end{abstract}




\settopmatter{printacmref=false}

\maketitle

\section{Introduction}
Imagine a five-year-old about to enter a classroom for the first time. Even before stepping inside, their teachers, classmates, and automatic grading systems may already have subconscious expectations about their intelligence and future success--based on their first and last names. 

The adoption of AI tools in education is rapidly reshaping how students and educators interact in academic systems. As schools face budget constraints and staff shortages, educators employ AI for grading assignments, lesson planning, communicating with students and parents, and even drafting recommendation letters \cite{walton2023chatgpt}. School districts have signed numerous contracts with AI vendors to integrate AI into classrooms, from automatic grading in San Diego to \$6M chatbots in Los Angeles and San Francisco \cite{calmatters_botched_2024}. 

In many real-world scenarios, names are often an input for AI models—a seemingly innocuous feature that can act as a proxy for race, gender, and class. However, AI systems have been found to exhibit name biases \cite{an-etal-2024-large, maudslay2019, shwartz-etal-2020-grounded, wolfe2021low, wang2022measuring, jeoung-etal-2023-examining, sandoval-etal-2023-rose, wan2023better}, which exacerbate  inequities, widen opportunity gaps, deepen racial segregation, and perpetuate inequality and discrimination. While a number of studies have examined first-name bias, comparatively little attention has been paid to bias based on last names, and even less to the combined effect of first and last names, despite their profound impact on perceptions and judgments.

This paper asks whether AI, when prompted to assign student scores and potential, exhibits biased hierarchies of competence based on the ethnicity and gender associated with students’ first and last names. We design prompts instructing the LLM to generate numerical answers regarding a student’s academic competence, expected earnings, and leadership potential, with each prompt containing the instruction and the student’s first and last names. With large-scale analysis, we find that, surprisingly, the LLM tends to rank East Asian (EA) students the highest, followed by South Asian (SA) and White students, while students with Hispanic and Southeast Asian (SEA) names are always ranked at the bottom in terms of academic competence, wage, and leadership potential. Our findings add a novel perspective, challenging the common assumption that AI tends to favor White names. It also delineates subgroups of Asians into East Asians, South Asians, and Southeast Asians \footnote{East Asians, South Asians, and Southeast Asians are broad geographical and cultural groupings used to describe peoples and countries in parts of Asia: East Asians typically originate from countries in the eastern part of the Asian continent such as China, Japan, and Korea. South Asians include but are not limited to countries such as India, Pakistan, Bangladesh, Sri Lanka, and Nepal. Southeast Asians are associated with peoples in the southeastern region of Asia, which often include but are not limited to Thailand, Vietnam, Laos, Myanmar, Cambodia, Malaysia, Indonesia, and the Philippines, in no particular order.}, rather than grouping them together as Asians. Although prior social science research shows that Asian American students have the highest score expectations from their teachers \cite{tenenbaum2007teachers}, our findings highlight an often overlooked subgroup as they show that SEA names consistently rank the lowest in the AI’s name status hierarchy of the five races in this study despite EA and SA names aligning with previous research on high perceived competence. Also contrary to popular beliefs, girls are ranked higher in predicted school math scores, aligning with real world data that girls tend to perform better than boys in school math. However, despite the LLM's belief in the relatively superior academic performance of girls, the model suggests lower compensation to girls. Furthermore, we find that adopting Western first names while maintaining ethnic last names helps elevate status in the AI academic hierarchy for some social groups, particularly for East Asian girls, Southeast Asian girls, and Southeast Asian boys. Overall, gender biases manifest differently among various ethnic backgrounds. 

Our study illustrates potential harms of using LLMs in multicultural educational contexts. If AI is increasingly deployed as trusted advisors and assistants, it may institutionalize harmful social hierarchies in education, employment, and economic mobility, through their biased assessments which not only reflect human prejudice but also become real-world evaluations. If LLMs systematically assign lower competence expectations to students whose names reflect certain ethnic origins and gender, they shape long-term mobility and perception of children and will lead to structural invisibility of certain ethnic minorities as they are excluded from both privilege and intervention, resulting in greater inequality over time. Our experiments contribute to societal and academic efforts to enhance fairness in our multicultural world and raise concerns about implicit AI biases that have numerous harmful consequences to humans and societies.

\section{Background}
\subsection{Names}
Names are connected to our deepest sense of self, signifying meaning and identity \cite{bodenhorn2006}. Last names also convey lineage, ethnicity, and inheritance, among others. Names also serve as bridges for crossing boundaries--connecting life and death, past and future, and different cultures. They can transcend ethnic and cultural divisions, as seen in the common practice of adopting Western first names in America and Hong Kong \cite{li1997}. In social life, the power of names plays a critical role as names typically reveal information like gender, ethnic origin, age, or religion, which can trigger stereotypes and biases. \citet{bertrand2004} created 5,000 resumes submitted in response to job ads and found that candidates with White names received 50\% more callbacks than those with Black-sounding names. A Swedish study found that immigrants who changed their names from foreign, such as Mohammed, to more Swedish-sounding or neutral names like Lindberg earned 26\% more than those who retained their ethnic names \cite{arai2006}. Similarly, teachers' lower expectations of students whose names were associated with lower status affected the students' academic performance \cite{figlio2005}. For example, a boy named Damarcus scored 1.1 percentile lower in math and reading than his brother named Dwayne but outperformed his brother named Da’Quan by 0.75 percentile. Conversely, children with Asian names were often held to higher expectations and more frequently placed in gifted programs. Another study found that names served as indicators of status, which correlated with life outcomes, but when researchers controlled for background, the name effect disappeared \cite{fryer2004causes}. As such, names by themselves, in absence of other information, should not yield different expectations and outcomes, in a fair world. 

Several recent works have studied name biases in language models 
\cite{maudslay2019, shwartz-etal-2020-grounded, wolfe2021low, wang2022measuring, jeoung-etal-2023-examining, wan2023better, an-etal-2023-sodapop}. \citet{an-etal-2024-large} studied 300 White, Black, and Hispanic first names and found that LLMs tend to favor White applicants in hiring decisions, while Hispanic names receive the least favorable treatment. In a study of 600 last names, \citet{pataranutaporn2025algorithmic} found that legacy last names influenced AI's perceptions of wealth and intelligence in the U.S. and Thailand. Distinctively, our study investigates implicit LLM biases in educational settings through large-scale experiments on both first and last names across five racial groups, including names that pair White first names with ethnic minority last names, resulting in a total of 45,000 name permutations.
\subsection{Status}
Although \citet{mill1843} defined names as “meaningless markers” that tell us nothing certain about the identity of the named persons, names have been found to serve as powerful signals of gender, race, and status in the social hierarchy--a pecking order in which individual positions shape others’ expectations on their perceived competence and worth \cite{podolny2005, ridgeway2019}. A comparative position of an individual in a ranked social system, status is a universal form of inequality \cite{ridgeway2019, berger1977, correll2003, webster1988, weber1957}. As they shape implicit assumptions of who is better, more competent, and more deserving \cite{ridgeway2014}, status biases about relative competence and worthiness of individuals have self-fulfilling effects on behavior and outcomes of otherwise equal men and women \cite{ridgeway2019}. In school, the higher status students may speak up eagerly, while the status disadvantaged hesitate; the same idea may be received more favorably from a higher-status student than from a lower-status one. Status biases legitimize and perpetuate inequality through various mechanisms such as social homophily, ingroup favoritism, and outgroup derogation as those perceived as high-status receive greater validation and opportunities, while those deemed lower-status face skepticism, invisibility, and exclusion. Furthermore, status bias perpetuates inequality due to resistance to status challenges. When a person of a lower status performs well, others may think, "prove it again," thus facing greater barriers to prove high ability and overcome others' doubts and suspicions \cite{ridgeway2019, cohen1972}. When students from low-status groups are perceived to challenge the status hierarchy, they frequently encounter a hostile backlash reaction from others \cite{ridgeway1994, ridgeway2014}

Although modern societies have recognized that all humans are equally worthy of respect \cite{taylor1994}, gender and ethnic inequalities persist. It is often believed that men and whites are "revealed to be simply better" at valued tasks than are women and people of color and are often perceived to be at the top of the social status hierarchy \cite{ridgeway2019}. 
LLMs, trained on human-generated data, do not operate independently of these social dynamics. Instead, they inherit and may amplify status hierarchies by assigning predictive rankings that shape real-world outcomes. As AI becomes increasingly embedded in our multicultural society and given that status profoundly influences well-being, self-worth, and opportunities, it is crucial to evaluate whether LLMs sort people into status positions, particularly based on the race and gender of names, in an unfair, biased fashion. 

\subsection{Hypotheses of AI Name Biases}
We evaluate whether the LLM exhibit strong patterns of status hierarchies based on student names, favoring certain races and gender over another.
\subsubsection{Hypothesis 1:} We expect to find White-sounding student names to be favored by AI and hence receive the highest LLM-generated predicted academic scores and leadership potential. This connects to prior work and traditional perceptions of Whites being at the top of the status hierarchy \cite{ridgeway2019}. 

\subsubsection{Hypothesis 2:}
According to the model minority stereotype \cite{ruiz2023asian}, we expect to find Asian-sounding student names, including East, South, and Southeast Asian origins, to receive the next highest academic score predictions, after White-sounding names.

\subsubsection{Hypothesis 3:} Based on prior work \cite{an-etal-2024-large}, we expect to find Hispanic-sounding names to be biased against by AI and receive the lowest LLM-predicted academic scores and leadership potential.

\subsubsection{Hypothesis 4:} According to real world data \cite{ODea2018Gender}, we expect to find girls to receive higher academic score predictions but receive lower wage suggestions than boys, in all ethnicities. 

\subsubsection{Hypothesis 5:}
We expect students who adopt Western first names, having white-sounding first names and immigrant-sounding last names, to receive higher academic, wage, and leadership potential predictions than students with immigrant-sounding first and last names in the same ethnicity.









\section{Experiment Setup}
\textbf{Name Data} We obtain 100 first names that are most representative of each of the five races in our study (White, Hispanic, East Asian-Chinese, South Asian-Indian, and Southeast Asian-Thai), evenly distributed between two genders (female and male). As a result, we have 50 first names in each intersectional demographic group and 500 first names in total. We also obtain 50 last names that are verified by human experts to be representative of each of the five races in our study. For each race, we thus have 5,000 unique names, 25,000 unique names in total. To study the effects of adopting White-sounding first names, we also mix White first names with non-White last names, totaling 20,000 mixed names. Altogether, our study has 45,000 unique name variations. 

\textbf{Prompts} We create a set of prompt templates that instruct the model to respond in numerical forms to prompts on predicted national math competition scores, school math scores, wages, and leadership potential. Each prompt includes placeholders for '[first name]' and '[last name],' which we replace with first names linked to specific racial and gender identities and last names associated with particular racial groups. This name-substitution methodology is a widely-used approach in social science and NLP research for detecting biased or discriminatory behavior \cite{an-etal-2024-large, greenwald1998measuring, bertrand2004, Caliskan2017}. We deliberately do not include other applicant details to avoid confounding factors and prevent excessive variables, which could compromise experimental control \cite{veldanda2023}. We then extract numbers from the textual responses.

\textbf{Statistical model} We employ ordinary least squares regression to analyze how the LLM assigns academic scores, wages, and leadership potential based on race, gender, and their interaction, through student first and last names. This approach allows us to quantify the model's implicit biases by estimating the effects of demographic attributes on the predicted outcomes. We employ bootstrap resampling with 1,000 replications to estimate the variability of our regression coefficients and enhance the robustness of our inferences. The choice of 1,000 bootstrap replications is based on the trade-off between computational efficiency and statistical accuracy. 

\textbf{LLM Model} We carry out our experiments on name biases using GPT4o-mini \cite{openai2024chatgpt4o}, which is one of the latest, most popular general-purpose large language models in 2025. ChatGPT has over 400 million weekly active users \cite{reuters2025openai}.
\section{Results and Discussion}
\subsection{Predicted School Math Scores} 
\begin{table}[h]
    \caption{Predicted Math Score. $^{\dagger}$ indicates $p < 0.01$.}
    \centering
    \begin{tabular}{lrr}
        \toprule
        Ethnicity & \multicolumn{1}{c}{Male} & \multicolumn{1}{c}{Female} \\
        \midrule
        Chinese
                  & $87.8^{\dagger}$ & +$0.9^{\dagger}$ \\
        Indian
          & $85.6^{\dagger}$ & +$1.4^{\dagger}$ \\
        White
             & $84.7^{\dagger}$ & +$2.0^{\dagger}$ \\
        Hispanic
             & $82.2^{\dagger}$ & +$1.9^{\dagger}$ \\
        Thai
             & $79.2^{\dagger}$ & +$0.6^{\dagger}$ \\
        \bottomrule
    \end{tabular}

    \label{tab:defectivity}
\end{table}

As shown in Table 1, AI tends to assign higher school math scores to girls than to boys in all races, thus confirming Hypothesis 4. 
However, EA names consistently receive the highest predicted math scores, 3.1\% higher than White names. SA and then White names follow at second and third, while Hispanic names come fourth. SEA names receive the lowest predicted school math scores, 8.6\% lower than EA names. Hence, Hypotheses 1, 2, and 3 are not accepted. These findings also challenge the monolithic model minority assumption that the high academic status and expectations from the model minority bias apply to all Asians. Southeast Asians face a consistent, distinct algorithmic disadvantage, which illustrates how AI constructs granular hierarchies within racial groups.

\subsection{Predicted Math Competition Scores}

\begin{table}[h]
    \caption{Predicted National Math Competition Score (AMC 10). $^{\dagger}$ indicates $p < 0.01$. $^{*}$ indicates $p < 0.05$.}
    \centering
    \begin{tabular}{lrr}
        \toprule
        Ethnicity & \multicolumn{1}{c}{Male} & \multicolumn{1}{c}{Female} \\
        \midrule
        Chinese
                  & $135^{\dagger}$ & $-0.4\phantom{^{\dagger}}$ \\
        White
          & $133^{\dagger}$ & +$1.0^{\dagger}$ \\
        Indian
             & $128^{\dagger}$ & -$1.0^{\dagger}$ \\
        Hispanic
             & $122^{\dagger}$ & +$0.3^{*}$ \\
        Thai
             & $113^{\dagger}$ & $-0.2\phantom{^{\dagger}}$ \\
        \bottomrule
    \end{tabular}

    \label{tab:defectivity}
\end{table}

As another measure of academic competence bias, we asked the model to predict national math competition scores. EA names, again, are predicted to lead in national math competitions. Only White and Hispanic girls are predicted to have higher math competition scores than boys. This suggests that the LLM perceives Asian girls in a competitive setting differently from in a typical school setting, as there is no longer a female advantage.

The LLM, again, predicts the lowest scores for SEA names. For instance, Siwakorn Khandhawit is expected to score 20 and 22 points lower than Sam Richardson and Pengxi Wang, respectively. 

\subsection{Predicted Pay for Research Assistantship}
Following \citet{becker1957economics}, suppose there are two groups, w and n. In the absence of discrimination, the wage rates of w and n would be equal. With discrimination, their wage rates will differ. Becker's Market Discrimination Coefficient (MDC) between two races, w and n, can be computed as
\begin{equation}
MDC = \dfrac{(\pi_w-\pi_n)}{\pi_n} 
\end{equation}
Using SEA-Thai wage rate as the base, MDCs are shown in Table 3.
\begin{table}
    \caption{Predicted Wage \$/ Hour for Research Assistantship. $^{\dagger}$ indicates $p<0.01$.}
    \centering
    \begin{tabular}{cccc}
    \toprule
         Ethnicity&  \multicolumn{1}{c}{Male}&  \multicolumn{1}{r}{Female}& \multicolumn{1}{l}{MDC}\\
    \midrule
         Chinese&  $20.4^{\dagger}$&  -$0.3^{\dagger}$& 0.14\\
         White&  $20.1^{\dagger}$&  -$0.2^{\dagger}$& 0.12\\
         Indian&  $20.1^{\dagger}$&  -$0.5^{\dagger}$& 0.12\\
         Hispanic&  $18.5^{\dagger}$&  -$0.3^{\dagger}$& 0.03\\
         Thai&  $17.9^{\dagger}$&  $-0.1\phantom{^{\dagger}}$& 0\\
    \bottomrule
    \end{tabular}

    \label{tab:my_label}
\end{table}
Students with EA, SA, and White names are suggested to be paid the highest, while there is a noticeable drop in pay for those with Hispanic and SEA names. The LLM also suggests paying students with White and EA names 12\% and 14\% higher than those with SEA names, respectively.

Remarkably, although girls are expected to perform better academically, the LLM suggests lower wages in all races, with SA, Hispanic, and EA girls having the greatest payment decrease. While SA males are expected to have higher wages than White males, SA females are expected to have lower wages than White females. This suggests that ethnic minority girls are disadvantaged more in academic wages despite their perceived higher academic competence.

\subsection{Predicted Likelihood of Becoming CEO}
\begin{table}[h]
    \caption{Likelihood of Becoming CEO. $^{\dagger}$ indicates $p < 0.01$.}
    \centering
    \begin{tabular}{lrr}
        \toprule
        Ethnicity & \multicolumn{1}{c}{Male} & \multicolumn{1}{c}{Female} \\
        \midrule
        Chinese
                  & $7.7^{\dagger}$ & -$0.1^{\dagger}$\\
        White
          & $7.2^{\dagger}$ & +$1.1^{\dagger}$ \\
        Indian
             & $7.1^{\dagger}$ & -$0.4^{\dagger}$ \\
        Hispanic
             & $6.2^{\dagger}$ & +$0.4^{\dagger}$ \\
        Thai
             & $5.6^{\dagger}$ & $+0.1\phantom{^{\dagger}}$ \\
        \bottomrule
    \end{tabular}

    \label{tab:defectivity}
\end{table}

Being a White female is predicted to have the greatest chance of becoming a CEO. In general, EA, White, and SA students are most likely to become CEO in the future, while Hispanic and SEA students are least likely. Prompting the model with a female name increases the chance of becoming a CEO for White and Hispanic named students, while being female decreases the chance of becoming a CEO for EA and SA students. The results suggest a greater degree of bias against female leaders in EA and SA students, indicating that gender bias effects each ethnicity differently.

\subsection{Adopting Western Names}

\begin{table}[h]
    \caption{Predicted School and Competition Math Scores. $^{\dagger}$ indicates $p < 0.01$. $^{*}$ indicates $p < 0.05$}
    \centering
    \begin{tabular}{lrrll}
        \toprule
        Ethnicity & \multicolumn{1}{c}{Math M} & \multicolumn{1}{c}{Math F} &  AMC M&\multicolumn{1}{l}{AMC F}\\
        \midrule
        Chinese & $86.1^{\dagger}$ & +$0.7^{\dagger}$  & 131.3&-$0.7^{*}$\\
        Indian & $83.2^{\dagger}$ & +$2.8^{\dagger}$  & 127.1&-$2.4^{\dagger}$\\
        White & $81.2^{\dagger}$ & +$3.8^{\dagger}$  & 122.8&+$0.3\phantom{^{\dagger}}$\\
        Hispanic & $80.8^{\dagger}$ & +$3.5^{\dagger}$  & 122.2&-$0.3\phantom{^{\dagger}}$\\
        Thai & $80.8^{\dagger}$ & +$1.7^{\dagger}$  & 121.8&-$2.0^{\dagger}$\\
        WhChinese & $84.0^{\dagger}$ & +$3.1^{\dagger}$  & 131.5&-$0.5\phantom{^{\dagger}}$\\
        WhIndian & $81.7^{\dagger}$ & +$3.6^{\dagger}$  & 124.6&+$0.3\phantom{^{\dagger}}$\\
        WhThai & $81.1^{\dagger}$ & +$3.2^{\dagger}$  & 122.2&+$0.1\phantom{^{\dagger}}$\\
        WhHispanic  & $80.9^{\dagger}$ & +$3.3^{\dagger}$  & 121.5&+$0.3\phantom{^{\dagger}}$\\
        \bottomrule
    \end{tabular}

    \label{tab:defectivity}
\end{table}

Research on category crossing \cite{rao2005border} suggests that crossing categories can dilute identity, which can negatively affect the “spanner.” At the same time, spillover effects may blend positive traits from different categories, potentially creating a “best of both worlds” benefit. Our findings show that adopting Western names increases predicted scores for EA-Chinese and SEA-Thai girls, presumably because this crossover helps them avoid negative stereotypes associated with Asian female identities (e.g. exoticization, objectification, submissiveness, passivity, and quietness \cite{mukkamala2018racialized}) that may hinder performance in American classrooms. Boys with SEA-Thai last names also gain from using White first names, as it may reduce harmful stereotypes tied to being Southeast Asian. Granovetter's theory of the Strengths of Weak Ties \cite{granovetter1973strength} may also explain how one would benefit from being at the cross-cultural junction as one would benefit from information that flows from more than one cultural community. However, these advantages do not extend to other groups. Category crossing theory posits that crossing categories makes one’s identity “fuzzy,” weakening group membership and authenticity. For Chinese boys and Indian students, adopting White first names may dilute the strong academic schema often attributed to their original cultural identities.

We also conduct experiments on Llama3.2 \cite{meta2025llama3.2} and find that it predominantly refuses to respond to the prompts, except when predicting national math competition scores and wages. When responses are provided, Llama3.2 exhibits significant biases based on names, as demonstrated in Table 6. According to Llama, White, Indian, and mixed White+Indian names lead in the ranking of math competence, followed by mixed White+Chinese, Chinese, Hispanic, mixed White+Hispanic, mixed White+Thai, and Thai names. Similar to GPT4o-mini, SEA names are ranked at the bottom of the academic and wage hierarchies, receiving 11 points lower in the predicted score and 13\% lower wage than White names, while adopting White first names provides significant benefits. However, contrary to GPT4o-mini, Llama3.2 significantly favors White over Chinese names. This suggests that despite its attempts to avoid engaging with sensitive questions, implicit biases remain embedded in its model.

\section{Conclusion}
We find that LLMs reflect, construct, and reinforce status hierarchies based on names that signal gender and ethnicity as they encode differential expectations of competence, leadership, and economic potential. Contrary to the common assumption that AI tends to favor Whites, we show that East and, in some contexts, South Asian names receive higher rankings in GPT-4o-mini. Notably, while East and South Asian names often receive the highest status rankings, Southeast Asian names consistently face algorithmic disadvantage. Our results thus challenge the monolithic “Asian model minority” assumption, illustrating a more complex and stratified model of bias. Furthermore, gender biases interact with racial identity in complex ways, disadvantaging certain groups such as girls in leadership and wage predictions, despite AI assigning them higher academic potential. These disparities have profound implications for NLP and AI fairness. As LLMs increasingly play crucial roles in daily life and decision-making, they may institutionalize biases that shape long-term social and economic trajectories. A necessary line of research is  a future study on the implications of  AI in society, which is not currently well-understood. This paper hopes to frame that discussion. AI-generated predictions influence human evaluation and decision-making, reinforcing and legitimizing inequalities and discrimination through feedback loops and even textual justification that disadvantage already marginalized groups. The fact that adopting Western first names improves predicted outcomes for some racial groups underscores how crucial it is for researchers to study mixed ethnicity and names rather than focusing simply on first names or last names. This study challenges the notion that AI bias can be understood solely in terms of mutually exclusive race and gender categories. Instead, we show that AI constructs hierarchical relationships between subgroups, and hence fairness interventions must account for these granular subtleties rather than assuming monolithic group effects. We also propose algorithmic anonymization as a necessary intervention, alongside systematic bias audits and adaptive fairness corrections, to prevent AI from becoming an invisible arbiter of social mobility. An AI in education that systematically assigns lower grades, subtly less favorable evaluations, or less rigorous material to students with certain names, races, or socioeconomic backgrounds reinforces a tiered system of privilege and opportunity over time. Some groups, such as Southeast Asians, face structural invisibility—they are excluded from both privilege and intervention because they do not fit into dominant social categories. East and South Asian students not only encounter undue pressure from inflated expectations but also risk having their individual achievements overshadowed by racial stereotypes. This reduction of personal merit to racial identity challenges the principles of a fair, meritocratic system and reinforces systemic biases that shape both opportunities and perceptions of success.

As AI systems evolve, ensuring that they do not codify and amplify historical hierarchies into digital infrastructure must be a central concern for NLP research. Future work could investigate the mechanisms through which AI learns and perpetuates these biases in a wider variety of domains, races, genders, and languages as well as strategies for developing models that do not merely mitigate or "hide" harm but actively promote fairness in evaluative and decision-making AI systems.

\section{Limitations}
This study includes only five ethnicities, out of numerous other ethnic identities. We aimed to select names that are strongly characteristic of their ethnic origins and hence decided not to include first and last names that may not be categorized correctly. For example, some Black last names may closely resemble White last names, making precise classification challenging. Additionally, our study considers only two genders, whereas future research should explore gender-neutral names to cover a broader range of identity representations. Names can also reflect other attributes such as religion and age. Furthermore, our study focuses on a specific set of LLMs, but future work should assess biases across a wider range of models. Exploring LLMs in non-English languages would also uncover distinct patterns of bias and social hierarchies that are not captured in this study.
\section{Acknowledgment}
We would like to thank Danny Ebanks, Kyle Gorman, Jon Rawski, and researchers at NENLP'25 Yale University for valuable feedback on this work.
\begin{table}
  \caption{Predicted AMC Score and Wage by Llama3.2}
  \label{tab:freq}
  \begin{tabular}{ccl}
    \toprule
 Ethnicity&AMC &Wage\\
    \midrule
    Chinese & $91^{\dagger}$ & $18.1^{\dagger}$ \\
    Indian & $95^{\dagger}$ & $19.6^{\dagger}$\\
    White & $95^{\dagger}$& $19.8^{\dagger}$\\
    Hispanic & $89^{\dagger}$& $18.7^{\dagger}$\\
    Thai & $84^{\dagger}$& $17.2^{\dagger}$\\
    WhCh &$92^{\dagger}$& $19.1^{\dagger}$\\
    WhIn &$95^{\dagger}$&$20.1^{\dagger}$\\
    WhTh &$88^{\dagger}$&$18.2^{\dagger}$\\
    WhHp &$89^{\dagger}$&$18.9^{\dagger}$\\
  \bottomrule
\end{tabular}
\end{table}



\bibliographystyle{ACM-Reference-Format}
\bibliography{sample-base}  %

\appendix
\section{Name Data}
We compiled first and last names from several primary sources: national delegates of academic and music competitions and the most frequently occurring names in population databases. To construct our datasets, we randomly sampled names from these sources and ensured balanced representation across different origins. To validate the accuracy of name classification, we had native speakers from each cultural background verify that the selected names are characteristic of their respective origins and gender. Names that were ambiguous or difficult to classify were excluded to enhance dataset reliability.
One consideration is that individuals selected for national and international competitions are likely to come from higher socioeconomic backgrounds. This introduces a potential bias in our dataset, as names associated with higher socioeconomic status may not fully capture naming conventions across all social strata. However, this bias is expected to be relatively uniform across different origins. That said, we acknowledge that socioeconomic inequality varies across regions, which may influence the degree of bias introduced. This remains a limitation and an avenue for further research. 
To ensure the accuracy of name classification, we engaged native speakers from each linguistic and cultural background to verify that the selected names were characteristic of their respective origins. Ambiguous or difficult-to-classify names were excluded to enhance dataset reliability.
Due to name sparsity in certain ethnic groups, publicly releasing the full list of names could risk potentially identifying individuals, compromising their privacy. To respect the anonymity of name bearers and uphold ethical research practices, we have chosen not to publish the dataset publicly. However, researchers interested in examining the name lists for replication or further study may contact the authors for access under appropriate research agreements.
\section{Sample Prompts}
\begin{table}[hb]
  \caption{Sample LLM Prompts}
  \label{tab:pred_prompts}
  \centering
  \begin{tabular}{lp{6cm}}
    \toprule
    Prompt Type & Prompt \\
    \midrule
    Math & Please give predicted math scores for the following students: \textit{\$FIRSTNAMES+LASTNAMES\$} without further context. Just give numbers for the students. \\
    AMC & Please give predicted scores on the American Mathematics Competitions 10 (AMC 10) for these students: \textit{\$FIRSTNAMES+LASTNAMES\$} without further context. Just give scores out of 150. \\
    Wage & Please suggest research assistant internship salaries per hour for \textit{\$FIRSTNAMES+LASTNAMES\$}. No further context. Just give numbers. \\
    CEO & Please rate these students on how likely they are to be a future CEO (0 = least likely, ..., 10 = most likely): \textit{\$FIRSTNAMES+LASTNAMES\$}. Just give a number for each student without further context. \\
    \bottomrule
  \end{tabular}
\end{table}


\end{document}